%% file: Near-Ideal.tex
\documentclass[11pt]{article}

\RequirePackage[OT1]{fontenc}
\RequirePackage{amsthm,amsmath}
\RequirePackage[numbers]{natbib}
\RequirePackage[colorlinks,citecolor=blue,urlcolor=blue]{hyperref}

\usepackage{amssymb,amsmath}
\usepackage{color}
\usepackage{graphicx}
\usepackage{tikz}

\input{notation.tex}

\input{colordef.tex}

\newcommand{\nmm}[1]{ \nm #1 \nm }
\newcommand{\nmeu}[1]{ \nm #1 \nm_2 }
\newcommand{\nmeusq}[1]{ \nm #1 \nm_2^2 }
\newcommand{\nmA}[1]{ \nm #1 \nm_A }
\newcommand{\nmP}[1]{ \nm #1 \nm_P }

\newcommand{\supp}{\mbox{supp}}
\def\xh{\hat{x}}

\def\xl{x_{\L}}

\def\xlo{x_{\L_0}}
\def\xloc{x_{\L_0^c}}
\def\hl{h_{\L}}
\def\hlc{h_{\L^c}}
\def\hlo{h_{\L_0}}
\def\hloc{h_{\L_0^c}}

\def\GkS{{\rm GkS}}
\def\ru{\underline{\r}}
\def\rb{\bar{\r}}

\def\nmsl1{\nm_{{\rm SL1}}}

{\bf}{\it}
\newtheorem{definition}{Definition}{\bf}{\it}
{\bf}{\rm}
\newtheorem{lemma}{Lemma}{\bf}{\it}
\newtheorem{theorem}{Theorem}{\bf}{\it}
{\bf}{\it}
{\bf}{\it}
{\bf}{\rm}

\newcommand{\argmin}{\operatornamewithlimits{argmin}}

\begin{document}

\title{
Near-Ideal Behavior of \\
Compressed Sensing Algorithms
}

\author{Eren Ahsen and M.\ Vidyasagar
\thanks{Erik Jonsson School of Engineering and Computer Science,
University of Texas at Dallas, Richardson, TX 75080.
Emails: \{ahsen,m.vidyasagar\}@utdallas.edu.
This research was Supported by the National Science Foundation under
Awards \#1001643 and \#1306630, and
the Cecil H.\ \& Ida Green Endowment at UT Dallas}
}

\maketitle






\begin{abstract}

In a recent paper \cite{Candes08}, see also \cite{DDEK12},
it is shown that the LASSO algorithm exhibits
``near-ideal behavior,'' in the following sense:
Suppose $y = Az + \eta$ where $A$ satisfies the restricted isometry
property (RIP) with a sufficiently small constant, and $\nm \eta \nm_2 \leq \e$.
Then minimizing $\nm z \nm_1$ subject to
$\nm y - Az \nm_2 \leq \e$
leads to an estimate $\xh$ whose error $\nm \xh - x \nm_2$ is bounded
by a universal constant times the error achieved by an ``oracle'' that
knows the location of the nonzero components of $x$.
In the world of optimization, the LASSO algorithm has been generalized in
several directions such as the group LASSO, the sparse group LASSO,
either without or with tree-structured
overlapping groups, and most recently,
the sorted LASSO.
In this paper, it is shown that {\it any algorithm\/} exhibits near-ideal
behavior in the above sense, provided only that (i) the norm used
to define the sparsity index is ``decomposable,'' (ii) the penalty
norm that is minimized in an effort to enforce sparsity
is ``$\g$-decomposable,'' and (iii) a ``compressibility condition''
in terms of a group restricted isometry property is satisfied.
Specifically, the group LASSO, and the sparse
group LASSO (with some permissible overlap in the groups), as well as the
sorted $\ell_1$-norm minimization all
exhibit near-ideal behavior.
Explicit bounds on the residual error are derived that contain
previously known results as special cases.

\end{abstract}


\section{Introduction}\label{sec:intro}

The field of ``compressed sensing'' has become very popular in recent years,
with an explosion in the number of papers.
In the interests of brevity, we refer the reader to two recent
papers \cite{DDEK12,NRWY12}, each of which contains an extensive bibliography.
Stated briefly, the core problem in compressed sensing is to approximate a
high-dimensional sparse (or nearly sparse) vector $x$ from a small number
of linear measurements of $x$.
Though this problem has a very long history (see the discussion in
\cite{DDEK12} for example), perhaps it is fair to say that much of
the recent excitement has arisen from \cite{Candes-Tao05},
in which it is shown that if $x$ has no more than $k$ nonzero components,
then by choosing the matrix $A$ to satisfy a condition known as the
restricted isometry property (RIP), it is possible to recover
$x$ exactly by minimizing $\nm z \nm_1$ subject to the constraint that
$Az = y = Ax$.
In other words, under suitable conditions,
among all the preimages of $y = Ax$ under $A$, the
preimage that has minimum $\ell_1$-norm is the sparse signal $x$ itself.
The same point is also made in \cite{Donoho06b}.
In case $y = Ax + \eta$ where $\eta$ is a measurement error and $x$
is either sparse or nearly sparse, one can attempt to recover $x$ by setting
\be\label{eq:11}
\xh := \argmin_{z \in \R^n} \nm z \nm_1 \st \nmeu { y - Az } \leq \e .
\ee
This algorithm is very closely related to 
the LASSO algorithm introduced in \cite{Tibshirani-Lasso}.
Specifically, the only difference between LASSO as in \cite{Tibshirani-Lasso}
and the problem stated above is that the roles of the objective function
and the constraint are reversed.
It is shown (see \cite[Theorem 1.2]{Candes08} that, under suitable
conditions, the residual error $\nmeu { \xh - x }$ satisfies
an estimate of the form
\be\label{eq:12}
\nmeu { \xh - x } \leq C_0 \s + C_2 \e ,
\ee
where $\s$ is the ``sparsity index'' of $x$ (defined below), and
$C_0,C_2$ are universal constants that depend only on the matrix $A$
but not $x$ or $\eta$.
The above bound includes exact signal recovery with noiseless measurements
as a special case,
and is referred to in \cite{Candes08} as ``noisy recovery.''

In a related paper \cite{Candes-Plan09}, the behavior of the
conventional LASSO algorithm is analyzed.
Thus one defines
\bd
\hat{\beta} := \argmin_{z \in \R^n} \nmeusq { y - Az } 
+ \l \nm z \nm_1 ,
\ed
where $\l$ is a Lagrange multiplier.\footnote{Note that there are
some slight departures from the notation used in \cite{Candes-Plan09}.
Specifically, in \cite{Candes-Plan09}
there a factor of $1/2$ in front of the $\ell_2$-norm,
and the Lagrange multiplier $\l$ is weighted by the standard deviation
of the measurement noise $\eta$.
But clearly these are very minor differences.}
The behavior of the estimate $\hat{\beta}$ is analyzed
under the assumption that
the vector $x$ is sparse, with the locations of the nonzero elements
chosen in accordance with a very nonrestrictive probabilitstic model.
It is shown that,
under suitable conditions on the matrix $A$ referred to as the ``coherence
property'' and for appropriate choices of the parameter $\l$,
the least squares error of the estimate $\xh$ produced by the
standard LASSO algorithm is ``nearly ideal'' in the following sense:
The residual error $\nmeusq { Ax - A \hat{\beta} }$
is bounded by a constant times the error of an ``oracle'' that knows
the locations of the nonzero components of $x$.

In the world of optimization, the LASSO algorithm has been generalized
in several directions, by modifying the $\ell_1$-norm penalty of
LASSO to some other norm that induces a prespecified
sparsity structure on the solution.
One such modification is
the Elastic Net (EN) algorithm introduced in \cite{Zou-Hastie05};
however the penalty in the EN algorithm is not a norm, though it is
a convex function.
Among the most popular sparsity-inducing penalty norms are
the group LASSO \cite{Yuan-Lin-Group-Lasso,Huang-Zhang10},
referred to hereafter as GL,
and the sparse group LASSO \cite{FHT10,SFHT12}, referred to hereafter as SGL.
Now there are versions of these algorithms that permit the groups
to overlap \cite{Jenetton-et-al11,OJV-Over-GL11}.
A recent contribution is to replace the usual $\ell_1$-norm
by the ``sorted'' $\ell_1$-norm \cite{Candes-sorted13}.
A similar idea is earlier proposed in \cite{Daubechies-et-al10},
but in the context of the sorted $\ell_2$-norm.

It is therefore natural to ask whether
inequalities analogous to (\ref{eq:12}) hold when the $\ell_1$-norm
in (\ref{eq:11}) is replaced by other sparsity-inducing norms such
as those mentioned in the previous paragraph.
More to the point, it would be highly desirable to have a general theory
of what properties a norm needs to satisfy, in order
that inequalities of the form (\ref{eq:12}) hold.
That is the focus of the present paper.
By a slight abuse of terminology, we will refer to the optimization
problem formulation in (\ref{eq:11}) as the ``LASSO algorithm,'' though
as mentioned earlier the roles of the objective function and constraint
are reversed in comparison to the conventional LASSO algorithm
of \cite{Tibshirani-Lasso}.
In the same vein, we will refer to the inequality (\ref{eq:22}) as
``near ideal behavior,'' and will interpret the results of
\cite{Candes08}, specifically \cite[Theorem 1.2]{Candes08}, as saying
that the LASSO algorithm exhibits near-ideal behavior.
Accordingly, the main objective of this paper is to present a very general
result to the effect that {\it any\/} compressed sensing algorithm
exhibits near-ideal behavior provided it satisfies three conditions:
\ben
\item A ``compressibility condition'' is satisfied, which
in the case of LASSO is that
the restricted isometry property (RIP) holds with a sufficiently
small constant.
\item The approximation
norm used to compute the sparsity index of the unknown vector
$x$ is ``decomposable'' as defined subsequently.
\item The penalty norm used to induce the sparsity of the solution, that is,
the norm that is minimized, is ``$\g$-decomposable'' as defined
subsequently.
\een
It will follow as a consequence of this general result that
sorted $\ell_1$-norm minimization, GL, and SGL (without  or with
tree-structured overlapping groups) all exhibit near-ideal
behavior.
In addition to the generality of the results established,
the method of proof is more direct than that in \cite{Candes08,DDEK12}.
Moreover, subject to one additional assumption that always holds in the
case of ``pure'' sparsity (but not in ``group'' sparsity), the results
presented here reduce to those in \cite{Candes08,DDEK12}.

In the literature on the behavior of various compressed sensing algorithms,
the measurement matrix $A$ is assumed to satisfy a variety of properties,
such the restricted isometry property (RIP), the coherence property, etc.
Almost all available methods for actually constructing matrices
satisfying these properties are {\it probabilistic}.\footnote{Note that
it is indeed possible to come up with
deterministic algorithms for constructing a measurement matrix with RIP.
However, the resulting matrix would be poorly conditioned from a
numerical standpoint.}
In other words, various methods are available which, {\it with high
probability}, result in a matrix $A$ having the desired properties.
As a result, the conclusions on near-ideal behavior are also probabilistic.
The paper \cite{DDEK12} , which is in turn an elaboration of \cite{Candes08},
is among the few to distinguish clearly between
the task of constructing a suitable measurement matrix, and the consequences
of doing so.
The former relies on probabilistic arguments whereas the latter relies
on purely deterministic arguments.
Accordingly, in this paper we adopt the same approach as in
\cite{Candes08,DDEK12}.
This demonstrates the versatility of the approach adopted in
\cite{Candes08,DDEK12}
to handle a variety of compressed sensing algorithms.

\section{Preliminaries}\label{sec:prelim}

As is by now customary in this area, if $x \in \R^n$, and $\L$ is a subset
of $\N = \{ 1 , \ldots , n \}$, the symbol $x_\L \in \R^n$ denotes
the vector such that $(x_\L)_i = x_i$ if $i \in \L$, and $(x_\L)_i = 0$
if $i \not \in \L$.
In other words, $\xl$ is obtained from $x$ by replacing $x_i$ by zero
whenever $i \not \in \L$.
Also, as is customary, for a vector $u \in \R^n$, its support set
is defined by
\bd
\supp(u) := \{ i : u_i \neq 0 \} .
\ed

Let $k$ be some integer that is fixed throughout the paper.
Next we introduce the notion of a group $k$-sparse set.
Some care is required in doing so, as the discussion following the
definition shows.

\begin{definition}\label{def:GKS}
Let $\G = \{ G_1 , \ldots , G_g\}$ be a partition of
$\N = \{ 1 , \ldots , n \}$, such that $| G_i | \leq k$ for all $i$.
If $S \seq \{ 1 , \ldots , g \}$, define $G_S := \cup_{i \in S} G_i$.
A subset $\L \seq \N$ is said to be {\bf $S$-group $k$-sparse}
for some $S \seq \{ 1 , \ldots , g \}$
if $\L = G_S$ and $| \L | \leq k$,
and {\bf group $k$-sparse} if
it is $S$-group $k$-sparse for some set $S \seq \{ 1 , \ldots , g \}$.
A vector $u \in \R^n$ is said to be {\bf group $k$-sparse}
if its support set $\supp(u)$ is contained in a group $k$-sparse set.
\end{definition}

At this point the reader might ask why a set $\L$ cannot be defined to
be group $k$-sparse if it is a {\it subset\/} of some $G_S$,
as opposed to being {\it exactly equal\/} to some $G_S$.
The reason is that, if every subset of $G_S$ is also called
``group $k$-sparse,''
then in effect all sets of cardinality $k$ or less can be called
group $k$-sparse, thus defeating the very purpose of the definition.
To see this, let $\L = \{ x_{i_1} , \ldots , x_{i_l} \}$, where $l \leq k$,
so that $| \L | = l \leq k$.
Then, since the sets $G_1 , \ldots , G_g$ partition the index set $\N$,
for each $j$ there exists a set $G_j$ such that $x_{i_j} \in G_j$.
Let $S \seq \{ 1 , \ldots , g \}$ denote the set consisting of all these
indices $j$.
Then $\L \seq G_S$.
So with this modified definition, there would be no difference between
group $k$-sparsity and conventional sparsity.
This is the reason for adopting the above definition.
On the other hand, it is easy to see that if $g = n$ and each set $G_i$
consists of exactly one element, then group $k$-sparsity reduces to
conventional $k$-sparsity.
Note also that a vector is defined to be group $k$-sparse if its
support {\it is contained in}, though not necessarily equal to,
a group $k$-sparse subset of $\N$.

Suppose $\nmm { \cdot } : \R^n \ap \R_+$ is some norm.
We introduce a couple of notions of decomposability that build upon
an earlier definition from \cite{NRWY12}.

\begin{definition}\label{def:decomp}
The norm $\nmm { \cdot }$ is said to be {\bf decomposable}
with respect to the partition $\G$ if, 
whenever $u, v \in \R^n$ are group $k$-sparse
with $\supp(u),\supp(v)$ contained in disjoint group $k$-sparse sets
$\L_u,\L_v$ respectively,
it is true that
\be\label{eq:21a}
\nmm { u + v } = \nmm { u } + \nmm { v } .
\ee
\end{definition}

As pointed out in \cite{NRWY12}, because $\nmm { \cdot }$ is a norm,
the relationship (\ref{eq:21a}) {\it always\/} holds with $\leq$ replacing
the equality.
Therefore the essence of decomposability is that the bound is tight
when the two summands are both group $k$-sparse vectors with their support
sets contained in disjoint group $k$-sparse subsets of $\N$.
Note that it is not required for (\ref{eq:21a}) to hold for every
pair of vectors with disjoint supports, only vectors whose support sets
are contained in disjoint group $k$-sparse subsets of $\N$.
For instance, if $\L$ is a group $k$-sparse set, and $u,v$
have disjoint support sets $\supp(u),\supp(v)$ that are both subsets of
$\L$, then there is no requirement that (\ref{eq:21a}) hold.
We will exploit this flexibility fully in Section \ref{sec:spec}
where it is shown that the penalty norms in the
group LASSO and sparse group LASSO algorithms can
be replaced by quite general norms and yet exhibit near-ideal behavior.

As shown in Section \ref{sec:spec}, the $\ell_1$-norm,
the group LASSO and the sparse group LASSO norm are all decomposable. 
However, the sorted $\ell_1$-norm is not decomposable.
To handle that case, we introduce a more general definition.

\begin{definition}\label{def:gdecomp}
The norm $\nmm { \cdot }$ is {\bf $\g$-decomposable} with respect to the
partition $\G$ if there exists $\g \in (0,1]$ such that the following is true:
whenever $u, v \in \R^n$ are group $k$-sparse
with $\supp(u),\supp(v)$ contained in disjoint group $k$-sparse sets
$\L_u,\L_v$ respectively, it is true that
\be\label{eq:21}
\nmm { u + v } \geq \nmm { u } + \g \nmm { v } .
\ee
\end{definition}

Note that if the norm $\nmm{\cdot}$ is $\g$-decomposable with
$\g=1$, then (\ref{eq:21}) and the triangle inequality imply that
\bd
\nmm { u + v } \geq \nmm { u } + \nmm { v } \implies \nmm { u + v }
= \nmm { u } + \nmm { v } .
\ed
Therefore decomposability is the same as $\g$-decomposability with $\g = 1$.

Clearly, if $\nmm { \cdot }$ is a decomposable norm, 
then (\ref{eq:21a}) can be applied recursively to show that
if $\L_0 , \L_1 , \ldots , \L_s$ are pairwise disjoint group $k$-sparse sets,
and $\supp(u_i) \seq \L_i$, then
\be\label{eq:22}
\nm \sum_{i=0}^s u_i \nm = \sum_{i=0}^s \nmm { u_i } ,
\ee
However, such an equality does not hold 
for $\g$-decomposable functions unless $\g=1$,
which makes the norm decomposable.
On the other hand, by repeated application of (\ref{eq:21})
and noting that $\g \leq 1$, we arrive at the following relationship:
if $\L_0 , \L_1 , \ldots , \L_s$ are pairwise group $k$-sparse sets,
and $\supp(u_i) \seq \L_i$, then
\be\label{eq:22b}
\nmm { \sum_{i=1}^s u_i } \geq \nmm { u_{\L_1} } 
+ \g \nmm { \sum_{i=2}^s u_i } .
\ee

Equation (\ref{eq:22}) is somewhat more general than the definition of
decomposability given in \cite{NRWY12},
in that we permit the partitioning of the index set $\N$ into more than
two subsets.
However, this is a rather minor generalization.\footnote{There is
a little bit of flexibility in \cite{NRWY12}
in that one can take two orthogonal subspaces that are not exactly
orthogonal complements of each other; but we will not belabor this point.}

With this preparation we can define the sparsity indices and optimal
decompositions.
Given an integer $k$, let $\GkS$ denote the collection of all group
$k$-sparse subsets of $\N = \{ 1 , \ldots , n \}$, and define
\be\label{eq:23}
\s_{k,\G} (x , \nmm { \cdot } ) := \min_{ \L \in \GkS} \nmm { x - \xl }
= \min_{ \L \in \GkS} \nmm { \xloc } 
\ee
to be the {\bf group $k$-sparsity index} of the vector $x$ with respect
to the norm $\nmm { \cdot }$ and the group structure $\G$.
Since the collection of sets $\GkS$ is finite (though it could be huge),
we are justified in writing $\min$ instead of $\inf$.
Once we have the definition of the sparsity index, it is natural to define
the next notion.
Given $x \in \R^n$, and a norm $\nmm { \cdot }$,
we call $\{ x_{\L_0} , x_{\L_1} , \ldots , x_{\L_s} \}$
an {\bf optimal group $k$-sparse decomposition} of $x$ if
$\L_i \in \GkS$ for $i = 0 , \ldots , s$, and in addition
\bd
\nmm { \xloc } = \min_{\L \in \GkS} \nmm { x - x_\L } ,
\nmm { x_{\L_i^c} } = \min_{\L \in \GkS} \nm x - \sum_{j=0}^{i-1} x_{\L_j} 
- x_\L \nm , i = 1 , \ldots , s .
\ed

\section{Problem Formulation}\label{sec:prob}

Throughout we shall make use of three distinct norms:
\bit
\item
$\nmA{ \cdot }$, which is a {\it decomposable norm\/} 
that is used to measure the quality of the approximation.
Thus, for a vector $x \in \R^n$, the quantity $\s_{k,\G} (x , \nmA { \cdot } )$
is the sparsity index used throughout.
\item
$\nmeu { \cdot }$, which is the standard Euclidean or $\ell_2$-norm, and
is used to constrain the measurement matrix via the group restricted
isometry property (GRIP).
\item
$\nmP { \cdot }$, which is a {\it $\g$-decomposable norm for some $\g \in (0,1]$\/}, 
that is minimized to induce a desired sparsity structure on the solution.
\eit
The prototypical problem formulation is this:
Suppose $x \in \R^n$ is an unknown vector,
$A \in \R^{m \times n}$ is a measurement matrix, 
$y = Ax + \eta$ is a possibly noise-corrupted measurement vector in $\R^m$,
and $\eta \in \R^m$ is the measurement error.
It is presumed that $\nmeu { \eta } \leq \e $, where $\e$ is a known
prior bound.
To estimate $x$ from $y$, we solve the following optimization problem
\be\label{eq:31}
\xh = \argmin_{z \in \R^n} \nmP { z } \st \nmeu { y - Az } \leq \e .
\ee

\begin{definition}\label{def:near-ideal}
The algorithm described in (\ref{eq:31}) for estimating $x$ is
said to be {\bf near ideal} or to show {\bf near-ideal behavior}, if
there exist universal constants\footnote{We use the symbols
$D_0$ and $D_2$, skipping $D_1$, to conform with the notation in
\cite{DDEK12}, where the corresponding constants are denoted by $C_0$ and
$C_2$.}
$D_0$ and $D_2$ that might depend on the matrix $A$ but not on
$x$ or $\eta$ such that\footnote{The symbol $A$ is unfortunately doing
double duty, representing the approximation norm as well as the measurement
matrix.
After contemplating various options, it was decided to stick to this
notation, in the hope that the context would make clear which usage is meant.}
\be\label{eq:32}
\nmeu{ \xh - x } \leq D_0 \s_{k,\G} ( x , \nmA { \cdot } ) + D_2 \e .
\ee
\end{definition}

The interpretation of the inequality (\ref{eq:32}) in this general setting 
is the same as in \cite{Candes08,DDEK12}.
Suppose the vector $x$ is group $k$-sparse, so that 
$\s_{k,\G} ( x , \nmA { \cdot } ) = 0$.
Then an ``oracle'' that knows the actual support set of $x$
can approximate $x$ through computing a generalized inverse of the columns
of $A$ corresponding to the support of $x$, and the resulting residual
error will be bounded by a multiple of $\e$.
Now suppose the algorithm satisfies (\ref{eq:32}).
Then (\ref{eq:32}) implies that the residual error
achieved by the algorithm is bounded by a universal constant times that 
achieved by an oracle.
Proceeding further, (\ref{eq:32}) also implies that if measurements
are noise-free so that $\e = 0$, then the estimate $\xh$ equals $x$.
In other words, the algorithm achieves exact recovery of group $k$-sparse
vectors under noise-free measurements.

The penalty norm $\nmP { \cdot }$ that is
minimized in order to determine a sparse approximation to $x$ need not
be the same as the approximation norm $\nmA { \cdot }$ used to measure
the quality of the resulting approximation.
This definition is motivated by the results in
\cite{Cohen-Dahmen-Devore09} in which
it shown that if $\nmP { \cdot }$ is the $\ell_1$-norm but $\nmA { \cdot }$
is the $\ell_2$-norm, then in essence compression is not possible.
Of course nothing prevents us from using the same norm for both penalization
and approximation.
Indeed, except for sorted $\ell_1$-norm minimization, we will take
both norms to be the same.

Regarding the use of the $\ell_2$-norm,
the theorems proved below are valid for any inner product norm, so
long as the partition is orthogonal with respect to the inner product. 
But the only results available in the literature for constructing
measurement matrices with the desired properties are for the
$\ell_2$-norm.
Hence in the current paper we will use $\ell_2$-norm to
constrain the measurement matrix $A$.

Throughout the paper, we shall be making use of four constants:
\be\label{eq:33}
a := \min_{x \neq 0} \frac{ \nmA { x} }
{ \nmP {x } } ,
b := \max_{\L \in \GkS} \max_{x_\L \neq 0} \frac{ \nmA { \xl } }
{ \nmP {\xl } } ,
\ee
\be\label{eq:34}
c := \min_{\L \in \GkS} \min_{x_\L \neq 0} \frac{ \nmA { \xl } }
{ \nmeu {\xl } } ,
d := \max_{\L \in \GkS} \max_{x_\L \neq 0} \frac{ \nmA { \xl } }
{ \nmeu {\xl } } .
\ee
Note that in the definition of $a$, the extremum is taken over all
nonzero vectors $x$, whereas in the definitions of $b, c, d$,
the extremum is taken only over all nonzero group $k$-sparse vectors.

Suppose for instance that $\nmA { \cdot } = \nmP { \cdot } = \nm \cdot \nm_1$,
which is the approximation as well as penalty norm used in LASSO.
Since $| \L | \leq k$ for all $\L \in \GkS$, we have by Schwarz's inequality that
\bd
\nm v \nm_1 \leq \sqrt{k} \nmeu { v }
\ed
whenever $\supp(v) \seq \L \in \GkS$.
In the other direction, we can write
\bd
v = \sum_{i \in \supp(v)} v_i \eb_i ,
\ed
where $\eb_i$ is the $i$-th unit vector.
Therefore by the triangle inequality
\bd
\nmeu { v } \leq \sum_{i \in \supp(v)} \nmeu { v_i \eb_i }
\leq \sum_{i \in \supp(v)} | v_i | = \nm v \nm_1 .
\ed
Therefore
\bd
1 \leq c \leq d \leq \sqrt{k} .
\ed
Estimates of these constants for other popular norms are given in
Section \ref{sec:spec}.

\section{A Known Result}\label{sec:known}

we begin by reprising a known result, as stated in \cite{DDEK12},
which is a more detailed description of ideas sketched in \cite{Candes08}.

\begin{definition}\label{def:RIP}
Suppose $A \in \R^{m \times n}$.
Then we say that $A$ satisfies the {\bf Restricted Isometry Property (RIP)}
of order $k$ with constant $\d_k$ if
\be\label{eq:41}
(1 - \d_k) \nm u \nm_2^2 \leq \langle u , Au \rangle \leq
(1 + \d_k) \nm u \nm_2^2 , \fa u \in \SI_k ,
\ee
where $\SI_k$ denotes the set of all $u \in \R^n$ such that
$| \supp(u) | \leq k$.
\end{definition}

\begin{theorem}\label{thm:41}
(\cite[Theorem 1.2]{Candes08}, \cite[Theorem 1.9]{DDEK12}; 
compare with \cite[Theorem 1.4]{Candes-Plan09})
Suppose $A \in \R^{m \times n}$ satisfies the RIP of order $2k$
with constant $\d_{2k} < \sqrt{2} - 1$, and that $y = Ax + \eta$ for some
$x \in \R^n$ and $\eta \in \R^m$ with $\nm \eta \nm_2 \leq \e$.
Define
\be\label{eq:42}
\xh = \argmin_{z \in \R^n} \nm z \nm_1 \st \nmeu { y - Az } \leq \e .
\ee
Then
\be\label{eq:42}
\nm \xh - x \nm_2 \leq C_0 \frac{ \s_k(x, \nm \cdot \nm_1) }{ \sqrt{k} }
+ C_2 \e ,
\ee
where
\be\label{eq:44}
C_0
= 2 \frac{ 1 + ( \sqrt{2} - 1 ) \d_{2k} } { 1 - ( \sqrt{2} + 1 ) \d_{2k} }
= 2 \frac{ 1 + \al } { 1 - \al } ,
\ee
\be\label{eq:45}
C_2
= \frac{ 4 \sqrt{ 1 + \d_{2k} } } { 1 - ( \sqrt{2} + 1 ) \d_{2k} }
= \frac { 4 \sqrt{1 + \d_{2k} } }  { (1 - \d_{2k} ) ( 1 - \al ) } ,
\ee
with
\be\label{eq:46}
\al = \frac{ \sqrt{2} \d_{2k} } { 1 - \d_{2k} } .
\ee
\end{theorem}

The formulas for $C_0,C_2$ are written slightly differently from those in
\cite[Theorem 1.9]{DDEK12} but are easily shown to be equivalent to them.

\section{Main Results}\label{sec:main}

In this section we present the two main theorems of the paper.
The first step is to extend the definition of RIP to group RIP.

\begin{definition}\label{def:GRIP}
A matrix $A \in \R^{m \times n}$ is said to {\bf satisfy the group RIP
of order $k$ with constants $\ru_k,\rb_k$} if
\be\label{eq:51}
0 < \ru_k \leq \min_{ \L \in \GkS} \min_{\supp(z) \seq \L} 
\frac{ \nm Az \nm_2^2 }{ \nm z \nm_2^2} ,
\ee
\be\label{eq:52}
\ru_k \geq \max_{ \L \in \GkS} \max_{\supp(z) \seq \L}
\frac{ \nm Az \nm_2^2 }{ \nm z \nm_2^2} .
\ee
We also define
\bd
\d_k := \frac{ \rb_k - \ru_k} {2} .
\ed
\end{definition}

Definition \ref{def:GRIP} shows that the group RIP constants $\ru_k$
and $\rb_k$ can be a lot closer together than the standard RIP constants
in Definition \ref{def:RIP}, because the various maxima and minima
are taken over only group $k$-sparse sets, and not all 
subsets of $\N$ of cardinality $k$.
This has implications when probabilistic methods are used to
construct the measurement matrix $A$.
We shall return to this topic in Section \ref{sec:sample}.

Note that in Definition \ref{def:RIP}, it was assumed that
the constants $\ru_k$ and $\rb_k$ are ``symmetric'' in that
they are of the form $1 - \d_k, 1 + \d_k$.
We could use a similar formulation for group RIP as well; but there
is no particular reason to do so.
The ``symmetric'' definition is used in Definition \ref{def:RIP} to 
facilitate comparison with earlier results, e.g.\ those in Theorem
\ref{thm:41}.

\begin{theorem}\label{thm:51}
Suppose the norm $\nmP { \cdot }$ is $\g$-decomposable, and that the
norm $\nmA { \cdot }$ is decomposable.
Define the constants $a, b, c, d$ as in (\ref{eq:33}) and (\ref{eq:34}).
Suppose $A \in \R^{m \times n}$ satisfies the group RIP property
of order $2k$ with constants $( \ru_{2k}, \rb_{2k} )$ respectively,
and let $\d_{2k} = ( \rb_{2k} - \ru_{2k} )/2$ as before.
Suppose $x \in \R^n$ and that $y = Ax + \eta$ where $\nmeu { \eta } \leq \e$.
Define
\be\label{eq:53}
\xh = \argmin_{z \in \R^n} \nmP { z } \st \nmeu { y - Az } \leq \e .
\ee
Suppose the compressibility condition
\be\label{eq:54}
\d_{2k} < \frac { a \g c \ru_k } { b d} .
\ee
is satisfied.
Then 
\be\label{eq:55}
\nmA { \xh - x} \leq \frac{r_3(r_2+1)}{1-r_1 r_2} \s_A
+ \frac{r_4(r_1+1)}{1-r_1 r_2} \e,
\ee
\be\label{eq:55a}
\nmeu { \xh - x} \leq \frac{r_3(r_2+1)}{a(1-r_1r_2)} \s_A + \frac{r_4(r_1+1)} 
{ a (1-r_1 r_2 ) }\e ,
\ee
where
\be\label{eq:57}
r_1=\frac{b}{a \g},r_2=\frac{ \d_{2k} d } {c \ru_k},
r_3=\frac{b(\g+1)}{a\g},r_4 =\frac {2d \sqrt{\rb_k} } { \ru_k },
\ee
\be\label{eq:57a}
\s_A := \s_{k,\G}(x , \nmA { \cdot } ).
\ee
\end{theorem}

Before giving the proof of the theorem, we make several observations.
Note that, unlike in the earlier result \cite[Theorem 1.9]{DDEK12},
stated here as Theorem \ref{thm:41}, the ``compressibility condition''
(\ref{eq:54}) involves $\ru_k$, and not $\ru_{2k}$.
Because the set of group $k$-sparse vectors is a subset of the set of
group $2k$-sparse vectors, it is obvious that
$\ru_{2k} \leq \ru_k \leq \rb_k \leq \rb_{2k}$.
So in this sense (\ref{eq:54}) is less conservative than any condition
that involves replacing $\ru_k$ by $\ru_{2k}$.

However, in the special case of RIP as opposed to group RIP,
a simplification is possible that does not work in more general situations.
Specifically, suppose $x \in \R^n$ and that $\{ x_{\L_0} , x_{\L_1} ,
\ldots , x_{\L_s} \}$ is an optimal $k$-sparse ({\it not\/} optimal
{\it group\/} $k$-sparse) decomposition of $x$
with respect to $\nm \cdot \nm_1$.
Then $x_{\L_0}$ consists of the $k$ largest components of $x$ by
magnitude, $x_{\L_1}$ consists of the next $k$ largest, and so on.
One consequence of this is that
\bd
\min_j | (x_{\L_i})_j | \geq \max_j | (x_{\L_{i+1}})_j | , \fa i ,
\ed
where the minimum on the left is taken over only the nonzero components.
Therefore
\be\label{eq:58}
\nmeu { x_{\L_{i+1} } } \leq \sqrt{k} \nm x_{\L_{i+1} } \nm_\infty
\leq \frac{1}{ \sqrt{k} } \nm x_{\L_i} \nm_1 .
\ee
This is the equation just above \cite[Equation(10)]{Candes08}.
However, when we take optimal {\it group\/} $k$-sparse decompositions, this
inequality is no longer valid.
For example, suppose $\nmP { \cdot } = \nm \cdot \nm_1$, 
let $n = 4 , g = 2 , k = 2$ and
\bd
G_1 = \{ 1 , 2 \} , G_2 = \{ 3, 4 \} ,
x = [ \ba{cccc} 1 & 0.1 & 0.6 & 0.6 \ea ]^t .
\ed
Then it is easy to verify that $s = 2$, and
\bd
\L_0 = \{ 3, 4 \} = G_2 , \L_1 = \{ 1 , 2 \} = G_ 1 ,
\ed
\bd
x_{\L_0} = [ \ba{cccc} 0 & 0 & 0.6 & 0.6 \ea ]^t ,
x_{\L_1} = [ \ba{cccc} 1 & 0.1 & 0 & 0 \ea ]^t .
\ed
Here we see that the largest element of $x_{\L_1}$ is in fact larger
than the smallest element of $x_{\L_0}$.
However, we do not have the freedom to ``swap'' these elements
as they belong to different sets $G_i$.
A more elaborate example is the following:
Let $n = 8, g = 4 , k = 4$, and
\bd
x = [ \ba{cccccccc} 0.1 & 1 & 0.2 & 0.3 & 0.4 & 0.5 & 0.4 & 0.7 \ea ] ,
\ed
\bd
G_1 = \{ 1 \} , G_2 = \{ 2, 3, 4 \} , G_3 = \{ 5, 6 \} , G_4 = \{ 7, 8 \} .
\ed
Then 
\bd
\L_0 = G_3 \cup G_4 , \L_1 = G_1 \cup G_2 .
\ed
Note that $x_{G_2}$ has higher $\ell_1$-norm than any other $x_{G_j}$.
However, since $G_2$ has cardinality $3$, it can only be paired with $G_1$,
and not with $G_3$ or $G_4$, in order that the cardinality of the union
remain less than $k = 4$.
And $\nm x_{G_1 \cup G_2} \nm_1 < \nm x_{G_3 \cup G_4} \nm_1$.

The proof of Theorem \ref{thm:51} depends on a few preliminary lemmas.

\begin{lemma}\label{lemma:51}
Suppose $h \in \R^n$, that $\L_0 \in GkS$ is arbitrary, and let
$h_{\L_1} , \ldots , h_{\L_s}$ be an optimal group $k$-sparse decomposition
of $\hloc$ with respect to the decomposable approximation norm $\nmA{ \cdot }$.
Then
\be\label{eq:59}
\sum_{j=1}^s \nmeu { h_{\L_j} } \leq \frac{1}{c} \nmA { \hloc } .
\ee
\end{lemma}

{\bf Proof:}
This is a direct consequence of the definition of the constant $c$
and the decomposability of $\nmA { \cdot }$.
We reason as follows:
\bd
\sum_{j=1}^s \nmeu { h_{\L_j} } \leq \frac{1}{c}
\sum_{j=1}^s \nmA { h_{\L_j} } = \frac{1}{c} \nmA { \hloc } .
\ed

Lemma \ref{lemma:51} is pretty straight-forward, and it is
not even necessary for $h_{\L_1} , \ldots , h_{\L_s}$ to be an
{\it optimal\/} group $k$-sparse decomposition -- it can be
{\it any\/} group $k$-sparse decomposition, as is easily verified.
But it is stated as a separate lemma just to facilitate comparison
with \cite[Equation (10)]{Candes08}, \cite[Lemma A.4]{DDEK12}.
Note that the summation in (\ref{eq:59}) begins with $j=1$ and not $j=2$
as in \cite{Candes08,DDEK12}.
If $\nmA { \cdot }$ is the $\ell_1$-norm, then $c = 1$, so the bound
in (\ref{eq:59}) is worse by a factor of $\sqrt{k}$
compared to that in \cite{Candes08,DDEK12}.
However, as is evident from the above example, in the case of
group $k$-sparse decompositions, in general it may not be possible
to do any better.

Lemma \ref{lemma:52} should be compared with 
\cite[Lemma 2.1]{Candes08}, \cite[Lemma A.3]{DDEK12}.

\begin{lemma}\label{lemma:52}
Suppose $A \in \R^{m \times n}$
satisfies the group RIP of order $2k$ with constants
$\ru_{2k}$ and $\rb_{2k}$, and that $u,v$ are group $k$-sparse
with supports contained in disjoint group $k$-sparse subsets of $\N$.
Then
\be\label{eq:510}
| \langle Au , Av \rangle | \leq \d_{2k} \nm u \nm_2 \cdot \nm v \nm_2 .
\ee
\end{lemma}

{\bf Proof:}
Since we can divide through by $\nm u \nm_2 \cdot \nm v \nm_2$,
an equivalent statement is the following:
If $u,v$ are group $k$-sparse with supports contained in disjoint
group $k$-sparse subsets of $\N$,
and $\nm u \nm_2 = \nm v \nm_2 = 1$, then
\bd
| \langle Au , Av \rangle | \leq \d_{2k} .
\ed
Now the assumptions guarantee that $u \pm v$ are both group $2k$-sparse.
Moreover $u^t v = 0$ since they have disjoint support.
Therefore $\nm u \pm v \nm_2^2 = 2$.
So the group RIP implies that
\bd
2 \ru_{2k} \leq \nm Au \pm Av \nm_2^2 \leq 2 \rb_{2k} .
\ed
Now the parallelogram identity implies that
\bd
| \langle Au , Av \rangle | =
\left| \frac{ \nm Au + Av \nm_2^2 - \nm Au - Av \nm_2^2 } {4} \right|
\leq \frac{ 2 ( \rb_{2k} - \ru_{2k} ) } {4} = \d_{2k} .
\ed

Lemma \ref{lemma:53} is the group analog of \cite[Lemma 1.3]{DDEK12},
which is also implicit in \cite{Candes08}.

\begin{lemma}\label{lemma:53}
Suppose $h \in \R^n$, that $\L_0 \in GkS$ is arbitrary, and let
$h_{\L_1} , \ldots , h_{\L_s}$ be an optimal group $k$-sparse decomposition
of $\hloc$ with respect to the decomposable approximation norm $\nmA{ \cdot }$.
Then
\be\label{eq:511}
\nmeu { \hlo } \leq \frac{ \d_{2k} } {c \ru_k} \nmA { \hloc } 
+ \frac { \sqrt{\rb_k} } { \ru_k } \nmeu { Ah } .
\ee
\end{lemma}

{\bf Proof:}
Note that $\hlo$ is group $k$-sparse.
Therefore by the definition of the group RIP property, it follows that
\bd
\ru_k \nmeusq { \hlo } \leq \nmeusq { A \hlo }
\leq \rb_k \nmeusq { \hlo } .
\ed
Next, observe that
\bd
\nmeusq { A \hlo } = \langle A \hlo , A \hlo \rangle .
\ed
So we will work on a bound for the right side.
We have that
\bd
\langle A \hlo , A \hlo \rangle
= \langle A \hlo , Ah \rangle - \langle A \hlo , A \hloc \rangle .
\ed
Next by Schwarz's inequality, it follows that
\beq
| \langle A \hlo , A \hloc \rangle |
& \leq & \left| \sum_{j=1}^s \langle A \hlo , A h_{\L_j} \rangle \right|
\nonumber \\
& \leq & \d_{2k} \nmeu { \hlo } \sum_{j=1}^s \nmeu { h_{\L_j} } 
\nonumber \\
& \leq & \frac{ \d_{2k} } {c} \nmeu { \hlo } \nmA { \hloc } . \nonumber
\eeq
Moreover
\bd
| \langle A \hlo , Ah \rangle | 
\leq \nmeu { A \hlo } \cdot \nmeu { A h } 
\leq \sqrt{ \rb_k } \nmeu { \hlo } \cdot \nmeu { A h } .
\ed
Combining everything gives
\beq
\ru_k \nmeusq { \hlo } & \leq & \nmeusq { A \hlo } \nonumber \\
& \leq & | \langle A \hlo , Ah \rangle | + | \langle A \hlo , A \hloc \rangle |
\nonumber \\
& \leq & \frac{ \d_{2k} } {c} \nmeu { \hlo } \nmA { \hloc }
+ \sqrt{ \rb_k } \nmeu { \hlo } \cdot \nmeu { A h } \nonumber .
\eeq
Dividing both sides by $\ru_k \nmeu { \hlo }$ leads to (\ref{eq:511}). $\qed$

{\bf Proof of Theorem \ref{thm:51}:}
Define $h := \xh - x$ and note that $x = \xh + h$.
Also, since $\xh$ is an optimizer as in (\ref{eq:53}), it follows that
$\nmP { x } \geq \nmP { \xh } = \nmP { x + h }$, which can be written as
\bd
\nmP { \xlo + \xloc } \geq \nmP { \xlo + \hlo  + \xloc + \hloc } .
\ed
The triangle inequality implies that 
\bd
\nmP { \xlo } + \nmP { \xloc } \geq \nmP { \xlo + \xloc } .
\ed
Combining this with the $\g$-decomposability of $\nmP { \cdot }$,
specifically (\ref{eq:22b}), we can reason as follows:
\beq
\nmP { \xlo } + \nmP { \xloc }
& \geq & \nmP { \xlo + \hlo } +\g \nmP { \xloc + \hloc } \nonumber \\
& \geq & \nmP { \xlo } - \nmP { \hlo } - \g \nmP { \xloc } + \g \nmP { \hloc } .
\nonumber
\eeq
Cancelling the common term $\nmP { \xlo }$ leads to
\be\label{eq:512}
\g \nmP { \hloc } \leq (\g+1) \nmP { \xloc } + \nmP { \hlo } .
\ee
By using the definitions of the constants $a,b$ from (\ref{eq:33}),
the triangle inequality, and 
the decomposability of the norm $\nmA{\cdot}$, we obtain
\beq
a \g \nmA { \hloc } & \leq & \g \nmP{\hloc}
\leq \nmP { \hlo } + (\g+1) \nmP { \xloc } \nonumber \\
& \leq & \nmP { \hlo } + (\g+1) \nmP{\sum_{j=1}^s x _{\L_j}} \nonumber \\
& \leq & b\nmA{ \hlo } +b(\g+1) \sum_{j=1}^s \nmA{ x _{\L_j}} \nonumber \\
& = & b \nmA{ \hlo } +b (\g+1) \nmA{\xloc}\nonumber\\
& = & b \nmA{ \hlo } +b (\g+1)  \s_A , \nonumber
\eeq
where $\s_A$ is used as a shorthand for $\s_{k, \G} ( x , \nmA { \cdot } )$.
Hence, dividing each side by $a \g$ and collecting all terms involving $h$
on the left side, we end up with 
\be \label{eq:513}
\nmA{\hloc} - r_1 \nmA{\hlo} \leq r_3 \s_A,
\ee
where $r_1$ and $r_3$ are given in (\ref{eq:57}).

We generate another inequality using (\ref{eq:511}), namely
\bd
\nmA { \hlo } \leq d \nmeu { \hlo }
\leq \frac{ \d_{2k} d } {c \ru_k} \nmA { \hloc }
+ \frac { d \sqrt{\rb_k} } { \ru_k } \nmeu { Ah } .
\ed
Since both $x$ and $\xh$ are feasible for the optimization
problem, we have that
\bd
\nmeu { Ah } = \nmeu { A \xh - A x } = \nmeu { y - A x - ( y - A \xh ) }
\leq 2 \e .
\ed
Substituting this, and collecting all terms involving $h$ on the left side,
leads to
\be\label{eq:513a}
- r_2 \nmA { \hloc } + \nmA { \hlo } \leq r_4 \e ,
\ee
where $r_2$ and $r_4$ are as in (\ref{eq:57}).

Equations (\ref{eq:513}) and (\ref{eq:513a}) can be together expressed as
a matrix inequality, namely:
\be\label{eq:514}
\left[ \ba{rr} 1 & -r_1 \\ -r_2 & 1 \ea \right]
\left[ \ba{c} \nmA{ \hloc } \\ \nmA { \hlo } \ea \right]
\leq \left[ \ba{c} r_3 \\ 0 \ea \right] \s_A
+ \left[ \ba{c} 0 \\ r_4 \ea \right] \e .
\ee
Let $M$ denote the $2 \times 2$ coefficient matrix on the left side.
It has positive diagonal elements and negative off-diagonal elements.
So it is easy to see that its inverse will be strictly positive if
and only if its determinant is positive.
Now the determinant of $M$ is positive if and only if $r_1r_2 <1$, which
is equivalent to (\ref{eq:54}).
Next, suppose (\ref{eq:54}) holds.
Then we can multiply both sides of (\ref{eq:514}) by $M^{-1}$ and get
\bd
\left[ \ba{c} \nmA{ \hloc } \\ \nmA{ \hlo } \ea \right]
\leq M^{-1} \left\{
\left[ \ba{c} r_3 \\ 0 \ea \right] \s_A
+ \left[ \ba{c} 0 \\ r_4 \ea \right] \e \right\} .
\ed
The triangle inequality now implies that
\beq
\nmA { h } & \leq & \nmA { \hloc } + \nmA { \hlo } \nonumber \\
& \leq & [ \ba{cc} 1 & 1 \ea ] \left[ \ba{c} \nmA{ \hloc } \\ \nmA { \hlo } \ea \right] \nonumber \\
& \leq &  [ \ba{cc} 1 & 1 \ea ] M^{-1} \left\{
\left[ \ba{c} r_3 \\ 0 \ea \right] \s_A
+ \left[ \ba{c} 0 \\ r_4 \ea \right] \e \right\} . \label{eq:515}
\eeq
Finally, observe that
\bd
M^{-1} = \frac{1}{1-r_1r_2} \left[ \ba{cc} 1 & r _1\\ r_2 & 1 \ea \right] .
\ed
Substituting into (\ref{eq:515}) and clearing terms leads to the
desired conclusion (\ref{eq:55}). Equation \eqref{eq:55a} 
follows from \eqref{eq:55} by using the definition of $a$.
$\qed$

To facilitate comparison with earlier results, we derive an alternate
version of Theorem \ref{thm:51}, based on an alternate version of
Lemma \ref{lemma:51}.
Suppose $h \in \R^n$, that $\L_0 \in GkS$ is arbitrary, and let
$h_{\L_1} , \ldots , h_{\L_s}$ be an optimal group $k$-sparse decomposition
of $\hloc$ with respect to the approximation norm $\nmA{ \cdot }$.
Suppose there exists a constant $f$ such that
\be\label{eq:516}
\sum_{j=2}^s \nmeu { h_{\L_j} } \leq \frac{1}{f} \nmA { \hloc } .
\ee
Note that the summation on the left side of (\ref{eq:516}) begins
with $j = 2$ and not $j= 1$ as in (\ref{eq:59}).
For instance, if we are studying conventional sparsity, wherein $g = n$
and each group consists of a singleton, it is known that $f$ can be
taken as $\sqrt{k}$.
See \cite[Equation(11)]{Candes08}, \cite[Lemma A.4]{DDEK12}.
With this assumption we can state another bound.

\begin{theorem}\label{thm:52}
Suppose $A \in \R^{m \times n}$ satisfies the group RIP property
of order $2k$ with constants $( \ru_{2k}, \rb_{2k} )$ respectively,
and let $\d_{2k} = ( \rb_{2k} - \ru_{2k} )/2$ as before.
Suppose $x \in \R^n$ and that $y = Ax + \eta$ where $\nmeu { \eta } \leq \e$.
Define
\be\label{eq:517}
\xh = \argmin_{z \in \R^n} \nmP { z } \st \nmeu { y - Az } \leq \e .
\ee
Suppose that
\be\label{eq:518}
\d_{2k} < \frac { f \ru_{2k} } { \sqrt{2} r_1d } .
\ee
Then
\be\label{eq:519}
\nmeu { \xh - x} \leq
\frac{ r_3 ( g + 1/f) }{ 1 - w } \s_A + \frac{ r_5 (1 + (r_1d)/f)}{ 1 - w } \e,
\ee
where
\be\label{eq:520}
g := \frac{ \sqrt{2} \d_{2k} }{ f \ru_{2k} } , w = r_1dg ,
r_5 := \frac{ 2 \sqrt{ \rb_k } } { \ru_k } ,
\ee
$\s_A$ are defined in (\ref{eq:57a}), $c$ is defined in (\ref{eq:34}).
\end{theorem}

If we compare the compressibility conditions (\ref{eq:518}) and
(\ref{eq:54}), we see that $\ru_k$ is replaced by $\ru_{2k}$ which is
in general smaller, and there is an extra factor of $\sqrt{2}$ in the
denominator.
Both of these are in favor of the bound in (\ref{eq:54}).
However, the constant $c$ is replaced by the constant $f$.
In the case of ``pure'' sparsity, $c = 1$ whereas $f = \sqrt{k}$, which
is a substantial advantage in favor of (\ref{eq:518}).
In the case of group sparsity, it is unclear whether
there is any advantage to (\ref{eq:518}).
As we shall see below, the main advantage of (\ref{eq:518}) and
of Theorem \ref{thm:52} is that it reduces to Theorem \ref{thm:41}
in the case of pure sparsity.

To prove the theorem, we need an alternate version of Lemma \ref{lemma:53},
which is entirely analogous to \cite[Lemma 1.3]{DDEK12}.

\begin{lemma}\label{lemma:54}
Suppose $h \in \R^n$, that $\L_0 \in GkS$ is arbitrary, and let
$h_{\L_1} , \ldots , h_{\L_s}$ be an optimal group $k$-sparse decomposition
of $\hloc$ with respect to the approximation norm $\nmA{ \cdot }$.
Define $\L = \L_0 \cup \L_1$.
Then
\be\label{eq:521}
\nmeu { \hl } \leq \frac{ \sqrt{2} \d_{2k} } {f \ru_{2k} } \nmA { \hloc }
+ \frac { \sqrt{\rb_{2k} } } { \ru_{2k}  } \nmeu { Ah } .
\ee
\end{lemma}

The proof is a fairly straight-forward modification of that of
Lemma \ref{lemma:53} and step by step the same as that of
\cite[Lemma 1.3]{DDEK12}.
But it is presented in detail, in the interest of completeness.

{\bf Proof of Lemma \ref{lemma:54}:}
Note that $\hl$ is group $2k$-sparse.
Therefore by the definition of the group RIP property, it follows that
\bd
\ru_{2k} \nmeusq { \hl } \leq \nmeusq { A \hl }
\leq \rb_{2k} \nmeusq { \hl } .
\ed
Next, observe that
\bd
\nmeusq { A \hl } = \langle A \hl , A \hl \rangle .
\ed
So we will work on a bound for the right side.
Note that
\bd
\langle A \hl , A \hl \rangle
= \langle A \hl , Ah \rangle - \langle A \hl , A \hlc \rangle .
\ed
Next by (\ref{eq:516}) and Schwarz's inequality, it follows that
\beq
| \langle A \hl , A \hlc \rangle |
& \leq & \left| \sum_{i=0}^1
\sum_{j=2}^s \langle A h_{\L_i} , A h_{\L_j} \rangle \right|
\nonumber \\
& \leq & \d_{2k} [ \nmeu { h_{\L_0} } + \nmeu { h_{\L_1} } ]
\sum_{j=2}^s \nmeu { h_{\L_j} }
\nonumber \\
& \leq & \frac{ \sqrt{2} \d_{2k} } {f} \nmeu { \hl } \nmA { \hlc } . \nonumber
\eeq
In the above, we use the known inequality
\bd
\nmeu { h_{\L_0} } + \nmeu { h_{\L_1} } 
\leq \sqrt{2} \nmeu { h_{\L_0} + h_{\L_1} } 
= \sqrt{2} \nmeu { \hl } ,
\ed
because $h_{\L_0}$ and $h_{\L_1}$ are orthogonal.
Next
\bd
| \langle A \hl , Ah \rangle |
\leq \nmeu { A \hl } \cdot \nmeu { A h }
\leq \sqrt{ \rb_{2k} } \nmeu { \hlo } \cdot \nmeu { A h } .
\ed
Combining everything gives
\beq
\ru_{2k} \nmeusq { \hl } & \leq & \nmeusq { A \hl } \nonumber \\
& \leq & | \langle A \hl , Ah \rangle | + | \langle A \hl , A \hlc \rangle |
\nonumber \\
& \leq & \frac{ \sqrt{2} \d_{2k} } {f} \nmeu { \hl } \nmA { \hlc }
+ \sqrt{ \rb_{2k} } \nmeu { \hl } \cdot \nmeu { A h } \nonumber .
\eeq
Dividing both sides by $\ru_{2k} \nmeu { \hl }$ leads to (\ref{eq:521}). $\qed$

{\bf Proof of Theorem \ref{thm:52}:}
As before, the optimality of $\xh$ and the $\g$-decomposability
of the penalty norm $\nmP{ \cdot }$ imply that
\bd
\nmA { \hloc } \leq r_3 \nmA { \xloc }+r_1 \nmA{ \hlo },
\ed
where $r_1$ and $r_3$ are given in (\ref{eq:57});
this is the same equation as (\ref{eq:513}).
Now rewrite this as
\be\label{eq:522}
\nmA { \hloc } 
\leq r_3 \s_A + r_1 \nmA { \hlo } \leq r_3 \s_A + r_1d \nmeu { \hlo } 
\leq r_3 \s_A + r_1d \nmeu { \hl } ,
\ee
because $\L_0$ is a subset of $\L$.
Next, the feasibility of both $x$ and $\xh$ implies that
$\nmeu { Ah } \leq 2 \e$.
Therefore (\ref{eq:521}) now becomes
\be\label{eq:523}
\nmeu { \hl } \leq \frac{ \sqrt{2} \d_{2k} } {f \ru_{2k} } \nmA { \hloc }
+ \frac { 2 \sqrt{\rb_{2k} } \e } { \ru_{2k}  } .
\ee
The inequalities (\ref{eq:522}) and (\ref{eq:523}) can be
written as a matrix inequality, namely
\bd
\left[ \ba{rr} 1 & -r_1d \\ -g & 1 \ea \right]
\left[ \ba{l} \nmA { \hloc } \\ \nmeu { \hl } \ea \right]
\leq \left[ \ba{c} r_3 \\ 0 \ea \right] \s_A
+ \left[ \ba{c} 0 \\ 1 \ea \right] r_5 \e ,
\ed
where $g$ and $r_5$ are defined in (\ref{eq:520}).
The coefficient matrix on the left side has a strictly positive inverse
if its determinant is positive.
Moreover, the determinant is just $1 - r_1dg = 1 - w$, where $w$
is also defined in (\ref{eq:520}).
So the compressibility condition is $w < 1$, which is clearly equivalent
to (\ref{eq:518}).
Moreover, if $w < 1$, then one can infer from the above matrix inequality that
\beq
\left[ \ba{l} \nmA { \hloc } \\ \nmeu { \hl } \ea \right]
& \leq & \frac{1}{ 1 - w } \left[ \ba{rr} 1 & r_1d \\ g & 1 \ea \right]
\left\{ \left[ \ba{c} r_3 \\ 0 \ea \right] \s_A
+ \left[ \ba{c} 0 \\ r_5 \ea \right] \e \right\} \nonumber \\
& = & \frac{1}{ 1 - w } \left\{ \left[ \ba{c} 1 \\ g \ea \right] r_3 \s_A
+ \left[ \ba{c} r_1d \\ 1 \ea \right] r_5 \e \right\} . \nonumber
\eeq
Now by (\ref{eq:516}),
\bd
\nmeu { \hlc } \leq \sum_{j=2}^s \nmeu { h_{\L_j} }
\leq \frac{1}{f} \nmA { \hloc } .
\ed
Therefore, since $h = \hl + \hlc$, the triangle inequality implies that
\beq
\nmeu { h } 
& \leq & \nmeu { \hlc } + \nmeu { \hl } \nonumber \\
& \leq & \frac{1}{f} \nmA { \hloc } + \nmeu { \hl } \nonumber \\
& \leq & \frac{1}{ 1 - w } [ \ba{cc} 1/f & 1 \ea ]
\left\{ \left[ \ba{c} 1 \\ g \ea \right] r_3 \s_A
+ \left[ \ba{c}r_1 d \\ 1 \ea \right] r_5 \e \right\}  \nonumber \\
& = & \frac{1}{ 1 - w } [ r_3 ( g + 1/f) \s_A + (1 + (r_1d)/f) r_5 \e ] . \nonumber
\eeq

\section{Special Cases}\label{sec:spec}

In this section it is shown that Theorem \ref{thm:51}
is general enough to encompass several algorithms such as
group LASSO, sparse group LASSO, either without overlapping groups
or with groups that overlap but have a tree structure, and 
sorted $\ell_1$-norm minimization.
Estimates for the constants $a$, $b$, $c$ and $d$ in (\ref{eq:33}),
(\ref{eq:34}) that appear in this theorem are derived for each of
these algorithms.
Then Theorem \ref{thm:52} is applied to pure sparsity, and 
Theorem \ref{thm:41} is derived as a special case.
Note that our use of terminology such as LASSO, group LASSO is
consistent with that in \cite{Candes08,DDEK12}, but inconsistent with
that in, for instance, \cite{Candes-Plan09}, in the sense that
the roles of the objective function and constraint are interchanged.

First we begin with various versions of the group LASSO algorithm.
As before, let $\G = \{ G_1 , \ldots , G_g \}$ be a partition of
the index set $\N = \{ 1 , \ldots , n \}$.
Let $\nm \cdot \nm_i : \R^{| G_i |} \ap \R_+$ be
{\it any\/} norm, and define the corresponding norm on $\R^n$ by
\bd
\nmA { x } = \sum_{i=1}^g \nm x_{G_i} \nm_i .
\ed
Then it is easy to see that the above norm is decomposable.
The exact nature of the individual norms $\nm \cdot \nm_i$ is
entirely irrelevant.
So if we define $\nmA { \cdot }$ as above and $\nmP { \cdot } = \nmA { \cdot }$,
then it follows from Theorem \ref{thm:51} that 
the corresponding algorithm exhibits near-ideal behavior,
provided an appropriate compressibility condition holds.

By defining the individual norms $\nm \cdot \nm_i$ appropriately,
it is possible to recover the group LASSO \cite{Yuan-Lin-Group-Lasso,
Huang-Zhang10}, the sparse group LASSO \cite{FHT10,SFHT12},
and the overlapping sparse group LASSO with tree-structured norms
\cite{Jenetton-et-al11,OJV-Over-GL11}.
The group LASSO norm is defined by
\be\label{eq:61}
\nm z \nm_{{\rm GL}} := \sum_{i=1}^g \nmeu { z_{G_i} } .
\ee
This corresponds to the choice $\nm \cdot \nm_i = \nmeu { \cdot }$.
Note that some authors use $\nmeu { z_{G_i} } / \sqrt{ | G_i | }$
instead of just $\nmeu { z_{G_i} }$.
The adjustments needed to handle to this variation are elementary
and are left to the reader.
The sparse group LASSO norm is defined by
\be\label{eq:62}
\nm z \nm_{{\rm SGL},\mu} := \sum_{i=1}^g
[ (1 - \mu) \nm z_{G_i} \nm_1 + \mu \nmeu { z_{G_i} } ] .
\ee
This corresponds to the choice
\bd
\nm z_{G_i} \nm_i = (1 - \mu) \nm z_{G_i} \nm_1 + \mu \nmeu { z_{G_i} } .
\ed

Next let us turn our attention to the case of ``overlapping'' groups
with tree structure,
as defined in \cite{Jenetton-et-al11,OJV-Over-GL11}.
In this case there are sets $\N_1 , \ldots , \N_l$, each of which
is a subset of $\N$, that satisfy the condition
\be\label{eq:63}
\N_i \cap \N_j \neq \es \implies ( \N_i \seq \N_j \mbox{ or }
\N_j \seq \N_i ) .
\ee
Though it is possible for some of these sets to overlap,
the condition above implies that the collection of sets $\N_1 , \ldots , \N_l$
can be renumbered with double indices as $\S_{ij}$,
and arranged in chains of the form
\bd
\S_{11} \seq \ldots \seq \S_{1n_1} , \ldots , 
\S_{s1} \seq \ldots \seq \S_{sn_s} ,
\ed
where the ``maximal'' sets $\S_{in_i}$ must also satisfy (\ref{eq:63}).
Therefore, given two maximal sets $\S_{in_i}, \S_{jn_j}$,
either they must be the same or they must be disjoint, because it is not
possible for one of them to be a subset of the other.
This shows that the maximal sets $\S_{in_i}$ are pairwise disjoint once
the duplicates are removed,
and together span the total feature set $\N = \{ 1 , \ldots , n \}$.
Thus, in a collection of tree-structured sets, the highest level sets
do not overlap!
Let $g$ denote the number of distinct maximal sets, and define
$G_i = \S_{in_i}$ for $i = 1 , \ldots, g$.
Then $\{ G_1 , \ldots , G_g \}$ is a partition of $\N$,
and each $\N_j$ is a subset of some $G_i$.
Now the authors of \cite{Jenetton-et-al11,OJV-Over-GL11} define
a combination of $\ell_1$- and $\ell_2$-norms on the various subsets
$\N_1 , \ldots , \N_l$ and add them up.
However, since each set $\N_j$ is a subset of some $G_i$,
a linear combination of norms on all sets $\N_j$
is just a sum of norms on the various $G_i$ and is therefore decomposable.
Indeed there is no reason to restrict oneself to convex combinations of
$\ell_1$- and $\ell_2$-norms on the subsets $\N_j$.
No matter what norms are defined on these subsets, the resulting
summation is still some linear combination of norms on the sets $G_i$,
and as a result the overall norm on $\R^n$ is decomposable.

Thus to summarize, the group LASSO norm, the sparse group LASSO norm, and
the penalty norms defined in 
\cite{Jenetton-et-al11,OJV-Over-GL11} are all decomposable.
Therefore the minimization of any of these norms leads to near-ideal behavior
under appropriate compressibility conditions.


Next we turn to the sorted $\ell_1$-norm introduced in \cite{Candes-sorted13},
which is defined as follows:
Suppose $\l_1 \geq \ldots \geq \l_n > 0$, and let $\bl \in \R_+^n$
denote the vector $[ \l_i , i = 1 , \ldots , n ]$.
Given a vector $x \in \R^n$, let $(x)$ denote its ``sorted'' version;
in other words, $(x)$ is a permutation of $x$, with the property that
$|(x)_i| \geq |(x)_j|$ if $i > j$.
Then the {\bf sorted $\ell_1$-norm} of $x$ is defined as
\be\label{eq:63}
\nm x \nmsl1 := \sum_{i=1}^n \l_i |(x)_i| .
\ee
It is not difficult to verify that the sorted $\ell_1$-norm is {\it not\/}
decomposable unless all $\l_i$ are equal, in which case it is
just (a multiple of) the $\ell_1$-norm.
However, it is $\g$-decomposable with $\g = \l_n/\l_1$.
To see this, suppose $u , v \in \R^n$ with disjoint supports,
and let $s_1 = | \supp(u) | , s_2 = | \supp(v) |$.
Since we have defined $\g$ as $\l_n/\l_1$, and
the $\l_i$ are nonincreasing, it is easy to see that
$\l_j / \l_i \geq \g$ whenever $1 \leq i < j \leq n$.
In particular we have that $\l_{s_1 + i} \geq \g \l_i$ for all $i$
between $1$ and $n - s_1$.
Also, since the vector $( u + v )$ consists of the elements of $u + v$
arranged in decreasing order of magnitude, and $u,v$ have
disjoint support sets, it is evident that
\bd
| ( u + v )_i | \geq | ( u + v )_j | \mbox{ if } i < j ,
\ed
\bd
| ( u + v )_i | \geq \max \{ | (u)_i | , | (v)_i | \} , i = 1 , \ldots ,
\min \{ s_1 , s_2 \} .
\ed
Therefore it follows from the definition of the sorted $\ell_1$-norm that
\beq
\nm u + v \nmsl1 & = & \sum_{i=1}^{s_1 + s_2} \l_i | ( u + v )_i | 
\nonumber \\
& = & \sum_{i=1}^{s_1} \l_i | ( u + v )_i |
+ \sum_{i=1}^{s_2} \l_{s_1 + i} | ( u + v )_{s_1 + i} | \nonumber \\
& \geq & \sum_{i=1}^{s_1} \l_i | ( u + v )_i |
+ \g \sum_{i=1}^{s_2} \l_i | ( u + v )_i | \nonumber \\
& \geq & \sum_{i=1}^{s_1} \l_i | (u)_i |
+ \g \sum_{i=1}^{s_2} \l_i | (v)_i | \nonumber \\
& = & \nm u \nmsl1 + \g \nm v \nmsl1 . \nonumber
\eeq
In the above line of reasoning, we have majorized $\l_{s_1 + i}$
by $\g \l_i$ and $| ( u + v )_{s_1 + i} |$ by $| ( u + v )_i |$.
The last step establishes that the sorted $\ell_1$-norm is $\g$-decomposable
with $\g = \l_n/\l_1$.
Therefore it follows from Theorem \ref{thm:51} that sorted $\ell_1$-norm
minimization leads to near-ideal behavior under appropriate compressibility
conditions.


In the remainder of the section, we
determine the constants $a$, $b$, $c$ and $d$ in (\ref{eq:33}), (\ref{eq:34})
that appear in Theorem \ref{thm:51} for various penalty norms.
We consider two situations:
In the first case, $\nmA { \cdot }$ is the group LASSO or sparse group LASSO
norm.
Since these norms are decomposable, we can as well take $\nmP { \cdot }
= \nmA { \cdot }$, which makes $a = b = 1$, and we need only to
determine the constants $c$ and $d$.
In the second case, $\nmA { \cdot }$ is the $\ell_1$-norm and
$\nmP { \cdot }$ is the sorted $\ell_1$-norm.

Consider now the group LASSO and the sparse group LASSO algorithms.
First, let $\nmA{\cdot} = \nmP { \cdot } = \nm \cdot  \nm_{{\rm GL}}$.
Then since $\nmP{\cdot}=\nmA{\cdot}$, we have $a = b = 1$.
To calculate $c$ and $d$, define $l_{{\rm min}}$ to be the smallest
cardinality of any $G_i$,
and define $s_{{\rm max}} := \lfloor k/ l_{{\rm min}} \rfloor$.
Now suppose that $\L \in \GkS$.
Specifically, suppose $\L = G_{i_1} \cup \ldots \cup G_{i_s}$.
Then clearly
\bd
\nm z_\L \nm_{{\rm GL}} = \sum_{j=1}^s \nmeu { z_{G_{i_j}} } ,
\ed
while
\bd
\nmeu { z_\L } = \left( \sum_{j=1}^s \nmeusq { z_{G_{i_j}} } \right)^{1/2} .
\ed
Thus, if we define the $s$-dimensional vector $v \in \R_+^s$ by
\bd
v = [ \nmeu { z_{G_{i_j}} } , j = 1 , \ldots , s ] ,
\ed
then
\bd
\nm z_\L \nm_{{\rm GL}} = \nm v \nm_1 , 
\nmeu { z_\L } = \nmeu { v } .
\ed
Now, by earlier discussion, we know that
\bd
\nmeu { v } \leq \nm v \nm_1 \leq \sqrt{s} \nmeu { v } .
\ed
Moreover, it is clear that the integer $s$, denoting the number of
distinct sets that make up $\L$, cannot exceed $s_{{\rm max}}$.
This shows that
\be\label{eq:64}
1 \leq c_{{\rm GL}} \leq d_{{\rm GL}} \leq \sqrt{s_{{\rm max}} } .
\ee

Next we analyze the sparse group LASSO algorithm, wherein
$\nmA { \cdot } = \nmP { \cdot } = \nm \cdot \nm_{{\rm SGL},\mu}$.
Then again we have that $a = b = 1$.
To calculate $c$ and $d$, suppose $\L = G_{i_1} \cup \ldots \cup G_{i_s}$.
Let $l_{{\rm max}}$ denote the largest cardinality of any $G_i$.
Then
\bd
\nmeu { z_{G_{i_j}} } \leq \nm z_{G_{i_j}} \nm_1 \leq
\sqrt{ l_{{\rm max}} } \nmeu { z_{G_{i_j}} } ,
\ed
whence
\be\label{eq:65}
\sum_{j=1}^s \nmeu { z_{G_{i_j}} } \leq \sum_{j=1}^s \nm z_{G_{i_j}} \nm_1 \leq
\sqrt{ l_{{\rm max}} } \sum_{j=1}^s \nmeu { z_{G_{i_j}} } .
\ee
Combining (\ref{eq:64}) and (\ref{eq:65}) leads to
\bd
\nmeu { z_\L } \leq \nm z_\L \nm_{{\rm SGL},\mu} \leq
[ ( 1 - \mu ) \sqrt{ l_{{\rm max}} } + \mu \sqrt{s_{{\rm max}} } ]
\nmeu { z_\L } .
\ed
Therefore
\be\label{eq:66}
1 \leq c_{{\rm SGL},\mu} \leq d_{{\rm SGL},\mu} 
\leq ( 1 - \mu ) \sqrt{ l_{{\rm max}} } + \mu \sqrt{s_{{\rm max}} } .
\ee

Finally consider the case where $\nmA { \cdot }$ is the $\ell_1$-norm and
$\nmP { \cdot }$ equals the sorted $\ell_1$-norm.
Then it is easy to see that
\bd
\l_n \nm x \nm_1 \leq \nmP { x } \leq \l_1 \nm x \nm_1 .
\ed
Therefore we can take
\bd
\frac{1}{\l_1} = a_{{\rm SL1}} \leq b_{{\rm SL1}} = \frac{1}{\l_n} .
\ed
It is already known that $1 \leq c \leq d \leq \sqrt{k}$.

We conclude this section by studying conventional (not group) $k$-sparsity,
and showing that the bounds in Theorem \ref{thm:41}
can be obtained from Theorem \ref{thm:52}.
First of all we have $\nmA{\cdot}=\nmP{\cdot}=\nm \cdot \nm_1$.
Since $\nmP{\cdot}$ is decomposable, we
have $\g=1$ and $\nmA{\cdot}=\nmP{\cdot}$ implies
that $a=b=1$. Therefore, $r_1=1$ and $r_3=2$,
where $r_1$ and $r_3$ are given in (\ref{eq:57}).
It is known from \cite[Equation (10)]{Candes08},
\cite[Lemma A.4]{DDEK12} that (\ref{eq:516}) holds
with $f = \sqrt{k}$.
since we are now dealing with conventional $\ell_1$- and
$\ell_2$-norms, we can take $c = 1$ and $d = \sqrt{k}$.
Moreover, the formulation of the RIP in (\ref{eq:41}) means that
\bd
\ru_{2k} = 1 - \d_{2k} , \rb_{2k} = 1 + \d_{2k} .
\ed
With these values, the constants $g$ and $w$ in (\ref{eq:520}) become
\bd
g = \frac{ \sqrt{k} \d_{2k} } { \sqrt{k} (1 - \d_{2k} ) } 
= \frac{ \al }{ \sqrt{k} } ,
w = r_1dg = \frac{ \sqrt{k} \d_{2k} } { (1 - \d_{2k} ) } = \al ,
\ed
where the constant $\al$ is defined in (\ref{eq:46}).
Therefore the two constants appearing in (\ref{eq:519}) become
\bd
\frac{ r_3 ( g + 1/f) }{ 1 - w } = \frac{2}{1 - \al}
\frac{1}{ \sqrt{k} } ( \al + 1 ) = \frac{2} { \sqrt{k} }
\frac{ 1 + \al }{ 1 - \al } = \frac{ C_0 }{ \sqrt{k} } ,
\ed
and
\bd
\frac{ r_5 (1 + (r_1d)/f)}{ 1 - w } =
\frac { 2 r_5 } { 1 - \al } = \frac{ 4 \sqrt{1 + \d_{2k} } }
{ ( 1 - \al ) ( 1 - \d_{2k} }  = C_2 ,
\ed
where $C_0,C_2$ are defined in (\ref{eq:44}) and (\ref{eq:45})
respectively.
Therefore we have rederived (\ref{eq:42}) using the bounds
in Theorem \ref{thm:52}, specifically (\ref{eq:519}).

\section{Sample Sizes}\label{sec:sample}

In this section we apply the method of \cite{BDDW08} to derive
the size of the measurement matrix in order for it to satisfy
the group RIP with a specified level of confidence.
Specifically, we consider the following problem:
Suppose the matrix $A$ is constructed by drawing $mn$ i.i.d.\ samples
of a random variable $\X$, so that $a_{ij} = \phi_{ij}(\om)$, where
$\om$ denotes the element of the probability space that generates 
the samples, and $i,j$ denote the row and column indices as usual.
Suppose that integers $n$ (the size of the vector $x$) and $k$
(the sparsity of $x$ though in unknown locations) are specified,
along with numbers  $\d$ denoting the group RIP constant and
$\zeta$ denoting a confidence parameter.
The objective is to derive lower bounds on the integer $m$ such that,
with probability no smaller than $1 - \zeta$, the $m \times n$ matrix $A$
satisfies the group RIP condition of order $k$ with constant no larger
than $\d$.

Specifically, suppose that $\X$ is a zero-mean random variable with
variance $1/m$, and let $\phi_{ij} (\om) , 1 \leq i \leq m , 1 \leq j \leq n$
be i.i.d.\ copies of $\X$, where $\om$ denotes the element of the sample space.
Let $\Phi ( \om ) \in \R^{m \times n}$ denote the resulting random
$m \times n$ matrix.
Then it is easy to verify that
\bd
E ( \nmeusq { \Phi ( \om ) v } ) = \nmeusq { v } .
\ed
Now suppose there exists a constant $c_0(\e)$ depending only on $\e$ such that
\be\label{eq:71}
\Pr \{ | \nmeusq { \Phi ( \om ) v } - \nmeusq { v } | > \e \}
\leq 2 \exp [ - m c_0(\e) ] .
\ee
Such bounds are established with $c_0(\e) = \e^2/4 - \e^3/6$
if $\X$ is Gaussian with zero mean and variance $1/m$ in \cite{BDDW08},
and for Bernoulli random variables in \cite{Achlioptas03}.

Now \cite[Lemma 6.1]{BDDW08} states that if $T \seq \N$ and $|T| \leq k$, then
\be\label{eq:73}
( 1 - \d ) \nmeusq { x_T } \leq \nmeusq { \Phi(\om) x_T }
\leq ( 1 + \d ) \nmeusq { x_T } , \fa x \in \R^n 
\ee
with probability $1 - \al'$, where
\bd
\al' \leq 2 \left( \frac{12}{\d} \right)^k e^{-m c_0(\d/2) } .
\ed
Now observe that, for Gaussian or Bernoulli variables,
\be\label{eq:72}
c_0 (\e) = \frac{ \e^2 }{4} - \frac{ \e^3} { 6} \geq \frac { \e^2 }{8} ,
\fa \e \leq 0.75 .
\ee
Therefore for Gaussian or Bernoulli variables, it follows that
\be\label{eq:74}
\al' \leq 2 \left( \frac{12}{\d} \right)^k e^{-m \d^2 / 32 } ,
\fa \d \leq 0.75 .
\ee

Equations (\ref{eq:73}) and (\ref{eq:74}) hold for {\it any one choice\/}
of the subset $T \seq \N$, no matter what it might be.
Therefore, if $\J \seq 2^\N$ is {\it any arbitrary collection\/} of
subsets of $\N$ where each set $T \in \J$ has cardinality $\leq k$,
then by the union bounds formula it follows that
\be\label{eq:75}
( 1 - \d ) \nmeusq { x_T } \leq \nmeusq { \Phi(\om) x_T }
\leq ( 1 + \d ) \nmeusq { x_T } , \fa T \in \J , \fa x \in \R^n
\ee
with probability $1 - \al$, where $\al = | \J | \al'$, that is
\be\label{eq:76}
\al \leq 2 | \J | \left( \frac{12}{\d} \right)^k e^{-m \d^2 / 32 }
\fa \d \leq 0.75 .
\ee
Therefore, given any collection of subsets $\J \seq 2^\N$ such that
each set $T \in \J$ has cardinality $\leq k$, all we have to do is to
ensure that the following inequality is satisfied:
\be\label{eq:77}
2 | \J | \left( \frac{12}{\d} \right)^k e^{-m \d^2 / 32 } \leq \zeta .
\ee
If every quantity is fixed except $m$, we can rewrite the above
inequality in terms of $m$, as follows:
\be\label{eq:78}
m \geq \frac{32}{\d^2} \left[ k \log \frac{12}{\d} + \log | \J |
+ \log \frac{2}{\zeta} \right] .
\ee

To apply this approach to constructing matrices that have the group RIP,
the only thing needed is to find an upper bound on the size of the
collection $\GkS$, the collection of all group $k$-sparse sets.
This is easily done using Sauer's lemma,\footnote{Actually Sauer's lemma
is more general than this, and bounds the number of subsets of a set
of cardinality $n$ that can be generated by intersecting with a collection
of VC-dimension $d$.}
which is well-known in
statistical learning theory, and states that for integers $1 \leq d \leq n$,
it is true that
\be\label{eq:79}
\sum_{i=0}^d \left( \ba{c} n \\ i \ea \right) \leq
\left( \frac{ e n }{d} \right)^d ,
\ee
where $e$ denotes the base of the natural logarithm.
Sauer's lemma can be found in many places, out of which 
\cite[Theorem 4.1]{MV-03} is just one reference.
To apply Sauer's lemma to the problem of group sparsity, as before let
$\G = \{ G_1 , \ldots , G_s \}$ denote a partition of $\N$.
Let $l_{{\rm min}}$ denote $\min_i | G_i |$, and let $s_{{\rm max}}$
denote the largest integer $s$ such that there exists a $\L \in \GkS$
of the form $\L = G_{i_1} \cup \ldots \cup G_{i_s}$.
Clearly $s_{{\rm max}} \leq k / l_{{\rm min}}$.
Therefore
\bd
| \GkS | \leq \sum_{i=0}^{s_{{\rm max}} } \left( \ba{c} g \\ i \ea \right)
\leq \left( \frac{ e g }{ s_{{\rm max}} } \right)^{s_{{\rm max}} } .
\ed
Therefore, given integers $n, m, k$ and numbers $\d \in (0,0.75],
\zeta \in (0,1)$, the measurement matrix $A = \Phi(\om)$ satisfies
the group RIP of order $k$ with coefficients $\ru_k \geq 1 - \d$
and $\rb_k \leq 1 + \d$, with probability at least $1 - \zeta$,
provided the following inequality holds,
which is obtained from (\ref{eq:78}) by replacing $| \J |$
by the bound from Sauer's lemma:
\be\label{eq:710}
m_{{\rm GS}} \geq \frac{32}{\d^2} \left[ k \log \frac{12}{\d}
+ s_{{\rm max}} ( 1 + \log g - \log s_{{\rm max}} )
+ \log \frac{2}{\zeta} \right] .
\ee
Here the subscript ``GS'' denotes ``group sparsity.''
On the other hand, for ``pure'' sparsity, $\GkS$ becomes the set of
all $k$-sparse subsets of $\N$, whose cardinality is bounded
(again by Sauer's lemma) by $(en/k)^k$.
Therefore
\be\label{eq:711}
m_{{\rm S}} \geq \frac{32}{\d^2} \left[ k \log \frac{12}{\d}
+ k ( 1 + \log n - \log k )
+ \log \frac{2}{\zeta} \right] .
\ee
Here the subscript ``S'' denotes ``sparsity.''

These formulas contain some good news and some bad news.
The first piece of good news is that the quantity $\zeta$
enters through the logarithm, so that
$m$ increases very slowly as we decrease $\zeta$.
This is consistent with the well-known maxim in statistical learning
theory that ``confidence is cheaper than accuracy.''
Another piece of good news is that asymptotically we have
\bd
m_{{\rm S}} \sim \frac{ k \log n }{\d^2} .
\ed
However, the {\it really bad news\/} is that in order for the 
logarithmic effect to start showing up, $n$ has to be really huge.

To illustrate this last comment, let us apply the bounds 
in (\ref{eq:710}) and (\ref{eq:711}) to typical numbers from
microarray experiments in cancer biology.
Accordingly, we take $n = 20,000$, which is roughly equal to the
number of genes in the human body and the number of measured quantities
in a typical experiment, and we take $k = 20$, which is a typical
number of key biomarkers that we hope will explain most of the observations.
Since $\d \leq \sqrt{2} - 1$ is the compressibility condition for
pure sparsity, we take $\d = 1/4 = 0.25$.
We partition the set of $20,000$ measurements into $g = 6,000$
sets representing the number of pathways that we wish to study,
and we take $l_{{\rm min}} = 4$, meaning that the shortest pathway
of interest has four genes.
Therefore we can take $s_{{\rm max}} = \lfloor k/ l_{{\rm min}} \rfloor = 5$.
Finally, let us take $\zeta = 10^{-6}$.
With these numbers, it is readily verified that
\bd
m_{{\rm S}} = 71,286 ,
m_{{\rm GS}} = 47,960 .
\ed
In other words, both values of $m$ are {\it larger than\/} $n$!
Therefore one can only conclude that the known bounds for $m$
are too coarse to be of practical use at least in computational biology,
though perhaps they might be of use in other applications where $n$ is
a few orders of magnitude larger.
For example, with $n = 10^6$, $k = 50$, $g = 3 \times 10^5$ and
$s = 5$, the choices $\d = 0.25$ and $\zeta = 10^{-6}$ lead to
\bd
m_{{\rm S}} = 385,660 ,
m_{{\rm GS}} = 137,260 .
\ed
Thus, though this time both $m_{{\rm S}}$ and $m_{{\rm GS}}$
are smaller than $n$, the difference is not substantial.

As to whether group sparsity offers any advantage over pure sparsity,
the evidence is not conclusive.
A comparison of (\ref{eq:710}) and (\ref{eq:711}) shows that
$m_{{\rm S}}$ grows as $( k \log n)/\d^2$, whereas 
$m_{{\rm GS}}$ grows as $( k + s_{{\rm max}} \log g)/\d^2$.
This observation is also made in \cite{Huang-Zhang10}.
Again, ``asymptotically'' the term involving $s_{{\rm max}}$ will dominate $k$
as $n$ becomes very large.
However, since $s_{{\rm max}}$ is multiplied by $\log g$,
$g$ would have to be truly enormous in order for group sparsity to lead
to substantially smaller values for $m$ than with pure sparsity.

The contents of this section can be summarized very simply.
While very general and universal bounds are available for RIP that
can be easily adapted to group RIP, these bounds are basically worthless
at the scale of numbers that appear in computational biology.

\section{Conclusions}\label{sec:concl}

In this paper we have presented a very general theory that allows us
to conclude that a compressed sensing algorithm shows near-ideal behavior
under extremely general conditions.
The results presented here show that not just LASSO, but also group LASSO,
sparse group LASSO, and their variants with tree-structured overlapping
groups, and sorted $\ell_1$-norm minimization, all show near-ideal behavior.
Along the way, we have introduced a notion of $\g$-decomposability,
which generalizes the notion of decomposability.
The next logical step would be to remove the restriction of tree structure
on overlapping groups.

\bibliographystyle{elsarticle-num}


\bibliography{Comp-Sens}

\end{document}

%% file: notation.tex
\def\eb{{\bf e}}

\def\G{{\cal G}}

\def\J{{\cal J}}

\def\N{{\cal N}}

\def\S{{\cal S}}

\def\X{{\cal X}}

\def\R{{\mathbb R}}

\def\al{\alpha}
\def\d{\delta}

\def\e{\epsilon}
\def\g{\gamma}

\def\l{\lambda}
\def\L{\Lambda}
\def\om{\omega}

\def\r{\rho}
\def\s{\sigma}
\def\SI{\Sigma}

\def\bl{{\boldsymbol \lambda}}

\def\ap{\rightarrow}

\def\es{\emptyset}
\def\seq{\subseteq}

\def\fa{\; \forall}

\def\st{\mbox{ s.t. }}

\def\nm{\Vert}

\renewcommand{\and}{\mbox{$\wedge$}}

\newcommand{\bc}{\begin{center}}
\newcommand{\ec}{\end{center}}
\newcommand{\be}{\begin{equation}}
\newcommand{\ee}{\end{equation}}
\newcommand{\bd}{\begin{displaymath}}
\newcommand{\ed}{\end{displaymath}}
\newcommand{\ba}{\begin{array}}
\newcommand{\ea}{\end{array}}
\newcommand{\ben}{\begin{enumerate}}
\newcommand{\een}{\end{enumerate}}
\newcommand{\bit}{\begin{itemize}}
\newcommand{\eit}{\end{itemize}}
\newcommand{\beq}{\begin{eqnarray}}
\newcommand{\eeq}{\end{eqnarray}}
\newcommand{\btab}{\begin{tabular}}
\newcommand{\etab}{\end{tabular}}
\newcommand{\bfig}{\begin{figure}}
\newcommand{\efig}{\end{figure}}
\newcommand{\btp}{\begin{tikzpicture}}
\newcommand{\etp}{\end{tikzpicture}}

%% file: colordef.tex
\definecolor{verm}{rgb}{0.6,0.2,0.2}
\definecolor{purp}{rgb}{0.3,0.1,0.6}
\definecolor{purple}{rgb}{0.4,0.0,0.6}
\definecolor{bggreen}{rgb}{0.1,0.3,0.1}
\definecolor{dgreen}{rgb}{0.1,0.6,0.1}
\definecolor{black}{rgb}{0.0,0.0,0.0}
\definecolor{crim}{rgb}{0.3,0.1,0.1}
\definecolor{dred}{rgb}{0.5,0.1,0.1}